# SCAPE: Searching Conceptual Architecture Prompts using Evolution

Soo Ling Lim
*Department of Computer Science*
*University College London*
London, United Kingdom
s.lim@cs.ucl.ac.uk

Peter J. Bentley
*Department of Computer Science*
*University College London*
*Autodesk Research*
London, United Kingdom
p.bentley@cs.ucl.ac.uk

Fuyuki Ishikawa
*National Institute of Informatics*
Tokyo, Japan
f-ishikawa@nii.ac.jp

*Abstract*—Conceptual architecture involves a highly creative exploration of novel ideas, often taken from other disciplines as architects consider radical new forms, materials, textures and colors for buildings. While today's generative AI systems can produce remarkable results, they lack the creativity demonstrated for decades by evolutionary algorithms. SCAPE, our proposed tool, combines evolutionary search with generative AI, enabling users to explore creative and good quality designs inspired by their initial input through a simple point and click interface. SCAPE injects randomness into generative AI, and enables memory, making use of the built-in language skills of GPT-4 to vary prompts via text-based mutation and crossover. We demonstrate that compared to DALL·E 3, SCAPE enables a 67% improvement in image novelty, plus improvements in quality and effectiveness of use; we show that in just three iterations SCAPE has a 24% image novelty increase enabling effective exploration, plus optimization of images by users. We use more than 20 independent architects to assess SCAPE, who provide markedly positive feedback.

## I. Introduction

Conceptual architecture is the discipline of expanding ideas within architectural design of buildings by the introduction of ideas from other disciplines. Eisenman [1] helped define the concept, with comparisons made between architecture and art in the creativity used by conceptual architects. Some conceptual architects explore abstract spaces and forms akin to the work of artists [2] as they expand the possibilities within architecture.

Given the proven ability of evolutionary computation (EC) to help generate creative designs [3, 4], the use of evolutionary algorithms within conceptual architecture is a well-established practice, dating back to the innovative work by Frazer [5] and the many works of Gero examining the interplay between creativity, evolution and architecture e.g., [6]. The related use of EC for the generation of art is also well-established, with the earliest demonstrations by Todd and Latham [7] and Sims [8] showing how powerful the evolutionary process can be.

Today there is a new type of generative system making headlines. Generative machine learning algorithms such as DALL·E, which make use of large language models and diffusion-based neural networks, are now hugely popular approaches for the generation of novel images [9]. From a text prompt these algorithms can generate high quality images of almost any subject in any style. As such they are now considered the current state of the art in image generation.

The quality of output from the latest generative AI models is unquestionable. But for the application of conceptual architecture, where extreme creativity is required, these models fall short. Trained to generate the most plausible output to match a text prompt, this means that for an architect to explore the space of possible ideas, they must first describe their idea in a prompt of sufficient detail, and then keep modifying that textual description. Prompt engineering is a challenging and increasingly complex task; finding exactly the right words to describe a subtle and ill-defined concept is infeasible for a creative explorer of architecture, especially when ideas may still be fluid and changing [10].

Generative AI models may be tuned to output more diverse results by maximizing the "temperature", but as this work will show, even when at their most "creative", these systems by design must produce outputs that closely match the most probable result for the text prompt. Because they are trained on millions of existing examples, they have little novelty or creativity of the type shown by evolution.

In contrast, evolutionary systems, which require no data and typically start from random and make use of randomized operators, are amazing innovators. Yet EC must use carefully designed representations, which often implicitly encode biases and constraints into the forms that evolution is able to generate [3]. EC can be creative, but cannot match the quality or realism of images produced by generative AI.

To address the needs of conceptual architecture, we introduce SCAPE (Searching Conceptual Architecture Prompts using Evolution) which combines methods from evolutionary computation with generative AI. We inject randomness into the generative AI system by embedding it within an evolutionary algorithm, using evolution to search "prompt space" on behalf of the user, and thus enable the best of both methods: the stunning realism of generative AI and the creativity of evolution. This work makes the following contributions:

- Novel combination of evolutionary computation with generative AI (DALL·E 3 and GPT-4).

- Novel use of evolutionary computation for human-in-the-loop LLM prompt evolution using textual attributes as genes.

- Novel use of LLMs for evolutionary initialization, mutation and crossover operators.

- Creation of a web application with interface designed to be easy-to-use for conceptual architecture.

- Comparison of SCAPE with vanilla DALL·E 3 in terms of ability to enable novelty, quality, effective use, exploration, and optimization, and with testing by real architects unaffiliated with this research.

## II. BACKGROUND

In previous work we combined machine learning (variational autoencoders) with evolution [11, 12] and showed it can be applied in practice to solve real-world problems [13]. Research into transformer architecture large language models (LLMs) such as GPT and diffusion-based DALL·E is considerable and rapidly advancing, as is their exploitation in commercial applications [14]. At the time of this work the latest version of GPT was 4 – rumoured[1] to be a mixture of experts model trained on much of the Internet with each expert fine-tuned for a specific type of query, e.g., generating code or text. It is believed that it comprises eight 222B parameter models. A version of GPT-4 that processes images is also available. Given an image, it produces detailed text describing the content of that image[2]. DALL·E 3 is the latest version of OpenAI's image generating model, trained on billions of images and associated text from the Internet. This work uses all three of these generative AI models as they are current state of the art, however it should be noted that the lack of scientific publications describing their internal workings, and reports that they may be being changed regularly behind the scenes mean that precise data on the models is subject to change.

Researchers have demonstrated that generative AI can be a useful tool in architecture [15, 16], and design in general, expanding capabilities from 2D to 3D [17]. Ideation in architecture has shown to be a viable application [18]. Millions of prompts were analysed to understand how users are already using such tools for architecture, with the conclusion that there is great potential [19], including for training in architecture [20]. The task of improving prompts attracts much work, with generative systems very sensitive to the precise wording used, resulting in the need for careful prompt engineering [21]. Methods of improving prompts using text modifiers [22] or using a human in the loop [23] are being developed. Evolution has been used with LLMs in many applications: e.g., for game level generation [24] and neural architecture search [25]. In a similar vein to our work, Yang et al. demonstrate how an LLM can optimize an objective function through prompt optimization [26], and Meyerson et al. demonstrate how LLMs can be used to perform crossover [27]. To date, however, the use of LLMs for initialization, crossover and mutation in prompt evolution as introduced in our work remains relatively unexplored.

## III. SCAPE

SCAPE operates like an evolutionary art system, where the user takes the role of the evaluation function. The user provides an initial text prompt comprising a few words; all subsequent input is provided by picking parents and optionally rating them in order to direct crossover and mutation operators, which construct the next generation of results. We make use of the LLM's knowledge of language to perform coherent crossover and mutation, akin to [27]. SCAPE is implemented as a web application, enabling it to be tested by anyone in the world. The backend is implemented in Django/Python, with MySQL as the database, and the frontend is implemented using JavaScript/jQuery, Bootstrap and HTML5. The web application was hosted on AWS Ubuntu Server 22.04 LTS (HVM). SCAPE uses GPT-4 (gpt-4-1106-preview, temperature=0.9, all other values default) and DALL·E 3 (dall-e-3, all values default). The source code and prompts are available on GitHub[3].

When designing the user interface, we considered the following elements. Our users may not be very technical, and they will want clean and beautiful interface, so we aim to make it self-explanatory and easy to use. We implement smooth scrolling so that their current required actions always show on top, while they can still scroll down to the bottom to see earlier images. We show all images in a row to enable comparison and enable zoom in for each image so that they can look at them in detail. API return from GPT-4 and DALL·E 3 is not in real time, so we added visual cues to let users know that computation is in progress.

### A. Representation

The genetic representation for SCAPE is text, where each individual in the population comprises: *initial_prompt* and six attributes: *architectural style, site, colors, lighting, shape/form, materials*.

Attributes (the genes to be evolved) are chosen to be relevant and commonly used by architects. The detailed operation of SCAPE follows below.

### B. Initialization

Once the web app is loaded, we take a short text input from the user to initialize the system, Fig. 1. To let the user understand SCAPE, we show the text: "SCAPE helps you explore novel ideas in conceptual architecture while improving aspects that matter to you," followed by a simple instruction for the user: "Start with a few words describing your architectural concept" with a placeholder of "A seashell inspired building." There is a minimum character limit of 4, i.e., the "Submit" button enables when there are more than 3 characters in the search bar.

Using the input provided by the user, we use GPT-4 to generate four individuals by generating values for the six attributes per individual. Our prompt to GPT-4 explicitly asks for the attribute values to be derived from the user text if possible and we define the required JSON format. We also specify that if no value can be extracted for a given attribute then random values appropriate for that attribute should be generated. By using GPT-4 in this way we make use of its ability to extract text, generate valid text, and produce correctly formatted output. While this is happening, user sees a "Generating attributes" spinner.

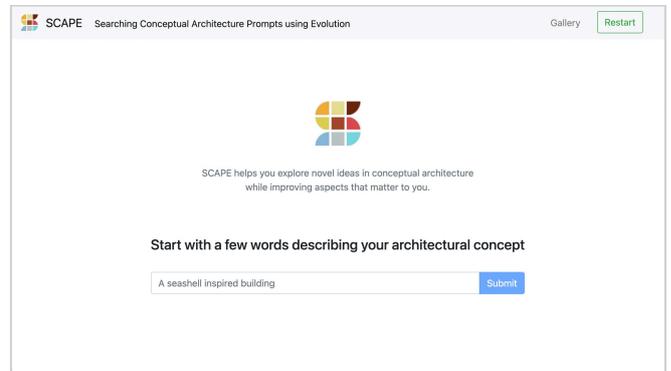

Fig. 1. SCAPE loading screen for initialization.

---

[1] https://pub.towardsai.net/gpt-4-8-models-in-one-the-secret-is-out-e3d16fd1eee0

[2] https://platform.openai.com/docs/guides/vision

[3] https://github.com/soolinglim/webllm/

TABLE I. INDIVIDUALS AND THEIR ATTRIBUTES

| Attributes | Individual 1 | Individual 2 | Individual 3 | Individual 4 |
|---|---|---|---|---|
| architectural style | Traditional Malay | Indigenous Bornean | Austronesian | Southeast Asian vernacular |
| colors | Natural wood, Red, Yellow | Earthy tones, Brown, Green | Ochre, Black, Natural timber | Dark wood, White, Blue |
| lighting | Natural daylight, oil lamps | Sunlight, torches | Firelight, daylight | Candles, natural light |
| materials | Wood, bamboo, thatch | Timber, rattan, leaves | Wood, thatched roofing | Bamboo, wood, palm leaves |
| shape/form | Rectangular, stilted | Long and elevated | Elongated, communal | Linear, multi-family |
| site | Rural village | Rainforest edge | Riverside | Near agricultural lands |

For clarity, we demonstrate this process in an example: a user input of "malaysian longhouse" results in four individuals with attributes shown in Table I.

### C. Image Generation

We call DALL·E 3 using a prompt to instruct it to behave as an architect and use the user prompt and attributes to generate an image suitable for an architecture magazine. This version of DALL·E 3 then uses GPT-4 to rewrite our prompt, generating a more complex written instruction for the generative image AI automatically. Once the image is generated, it is displayed on top of the attributes table, and the description generated by DALL·E 3 rewrite is displayed below the attributes table for the user. This is repeated for all four individuals using their corresponding attribute values. Once images are generated they are displayed, e.g., see Fig. 2.

### D. Parent Selection

We provide the user the ability to create the next generation of images, or to quit, with buttons "Improve these images" and "I'm done!" respectively. If the user clicks on the former, the user is then asked to select two best images as parents, and to optionally rate their attributes as good or bad. Users may pick images from any iteration by scrolling down in the window, i.e., parents may be selected from the current generation or any prior generation. When two images have been selected, the "Generate new images from feedback" button is enabled, Fig. 3.

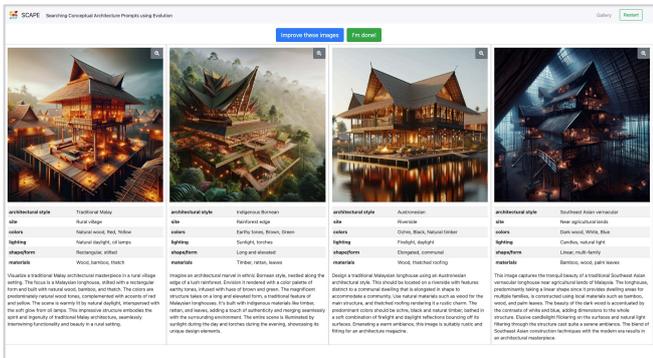

Fig. 2. Four generated images are displayed with their corresponding attributes and autogenerated description.

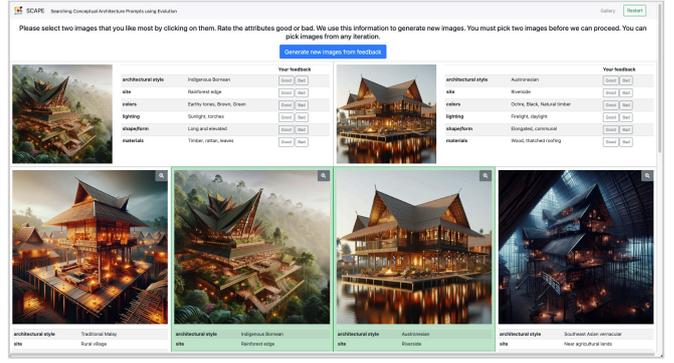

Fig. 3. Parent selection in SCAPE. Users may rate image attributes of the parents.

### E. Crossover and Mutation

Once the user clicks the "Generate new images from feedback" button, they see the "Generating attributes" spinner. In the background, the attribute ratings and parents are used for crossover and mutation to generate four new child solutions from the parents. The first stage of crossover is performed using GPT-4, with a prompt that specifies that a new value for each attribute should be generated that combines ideas from the values from each parent. (For example, attribute values of "Earthy tones with accents of electric lime" and "Chrome and electric blue" might result in a new crossover value of "Earth tones with chrome accents and electric blue highlights".) Then for each attribute, we construct a *choice list* by starting with [parent 1's value, parent 2's value, crossover generated value]. If the user has rated parent 1's value as "Good" we add two more parent 1's value into the list, so the list of choices becomes [parent 1's value, parent 1's value, parent 1's value, parent 2's value, crossover generated value]. We do the same for parent 2. The child's attribute value is then randomly chosen from the list.

To apply mutation, once crossover is complete and we have four child solutions, we then examine each child's attribute. We check to see if the current value of the attribute matches a value rated as bad for either parent 1 or parent 2 (where that match may be exact or similar, determined with another GPT-4 prompt, again making use of its knowledge of language). If it is a match then the attribute is added to the *attributes to mutate* list. If an attribute is unrated, then there is a 50% chance it will also be added to the *attributes to mutate* list (this is to allow more exploration, as users may not always rate something).

We then pass the list of *attributes to mutate* to GPT-4 (along with the list of all tabu values rated bad by the user) and specify that it should provide a novel alternative for each attribute that do not match the tabu values. For example, if *site* must not have the values ["Rural setting", "Coastal cliffside"], then the new random value might be "Mountainous highland". Again, we make use of GPT-4's knowledge of language to generate new alternatives, so we do not need dictionaries or look-up tables of possible words (a more restrictive and time-consuming method).

Once mutated, the new four child solutions are passed to DALL·E 3 for image generation as described above and the algorithm loops until the user clicks on "I'm done!"

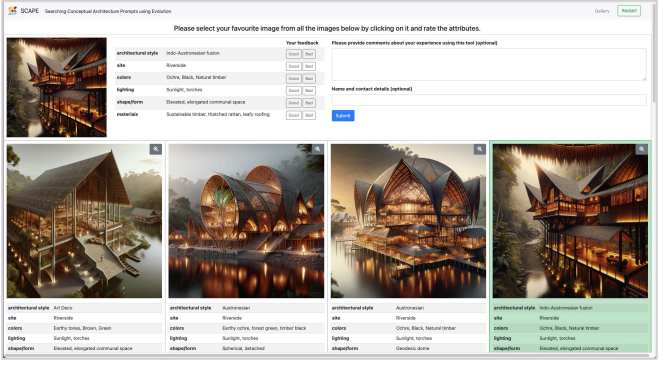

Fig. 4. Rating final best image and providing feedback.

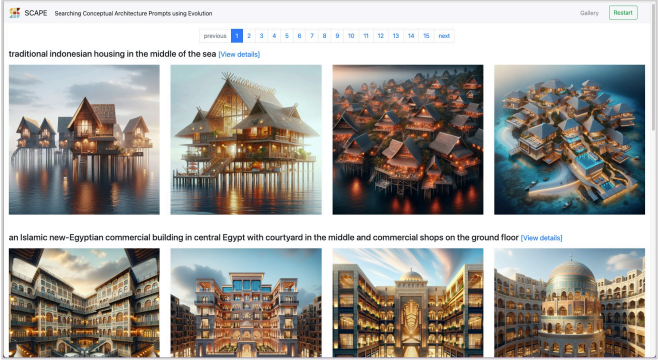

Fig. 5. SCAPE Gallery available to all users.

### F. Final Best Solution

When the user indicates they are finished, they will be prompted to select their favourite image from all the images below and to rate its attributes, see Fig. 4.

### G. Gallery of Evolved Images

SCAPE also provides a gallery of all images produced by the users to help users see the possibilities for the tool, Fig. 5.

### H. Justification of Design Decisions

There are many potential large generative models that could be used for this task. However, we found through initial experimentation that only relatively recent models are capable of reliably following the instructions and providing correctly formatted output, e.g., GPT-3.5 is not reliable enough, while GPT-4 works well; DALL·E 2 produces images of insufficient quality while DALL·E 3 is acceptable. Initially we provided examples for each attribute (when we tried GPT-3.5). With more advanced models we find zero-shot (no examples) works. This approach is advantageous because no bias is introduced. GPT is poor at producing random or different values – ask for random colors (without a history), and it always produces the same "random" color. This is why it is necessary to specify that it must avoid colors tried before or disliked by users.

## IV. EXPERIMENTS AND RESULTS

### A. Research Questions

We investigate the following questions in this work:

- RQ1: How does SCAPE compare to vanilla DALL·E 3? We compare using three metrics:
  1. Novelty: How good is SCAPE compared to vanilla DALL·E 3 in terms of providing novelty?
  2. Quality: How good is SCAPE at generating architecture from a short prompt?
  3. Effectiveness: How effectively can SCAPE find a desired concept?
- RQ2: How good is SCAPE at optimisation? We examine whether user satisfaction improves over time.
- RQ3: How good is SCAPE at exploration? We examine the diversity of designs produced over time.
- RQ4: How do real architects find the tool? We analyse usage of SCAPE and feedback by independent architects unaffiliated with this research.

### B. Experiment 1 - Comparing SCAPE to Vanilla DALL·E 3

#### 1) Novelty

We use pairwise comparison to compare the result from running DALL·E 3 four times with the first generation produced by SCAPE using the same prompt. To achieve consistent, scalable, and objective comparisons we use GPT-4V (gpt-4-vision-preview) and ask it to give us a 0-10 score for image difference, with output tuned to match human ratings through preliminary experiments. We repeat the experiment ten times using different prompts, taking the average score. Fig. 6 and Fig. 7 illustrate the results from this comparison for one prompt. We find SCAPE is able to produce 66.7% more diverse and novel images that contain different concepts as assessed by GPT-4V.

#### 2) Quality

We compare the output of SCAPE for the first iteration, and after user exploration, with the output from DALL·E 3 using the same prompt. Analysis of the images in terms of architectural plausibility (could this be considered a valid building architecture) and creativity (are there innovative concepts not obvious from the initial prompt) shows that SCAPE consistently outperforms vanilla DALL·E 3.

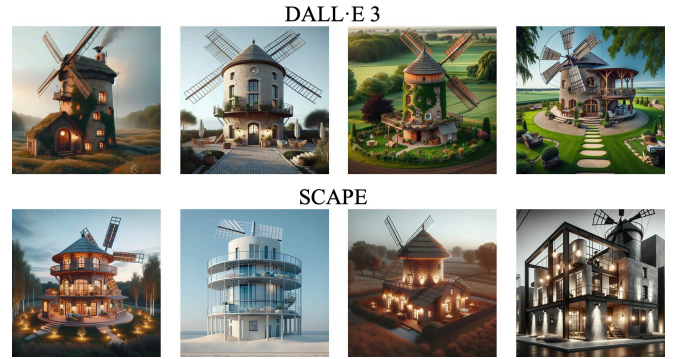

Fig. 6. Comparing novelty of image generation for prompt "windmill converted into dwelling".

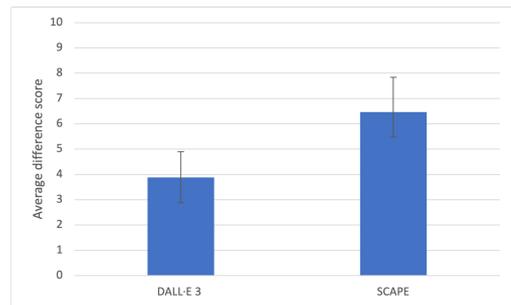

Fig. 7. Average difference score between sets of generated images reported by GPT-4V (higher = more different).

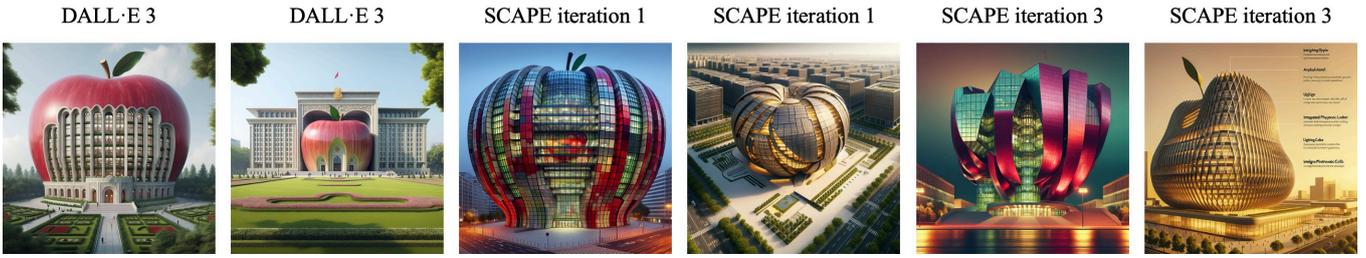

Fig. 8. Images produced by vanilla DALL·E 3 are architecturally naive compared to SCAPE iteration 1 and lack creativity compared to SCAPE iteration 3.

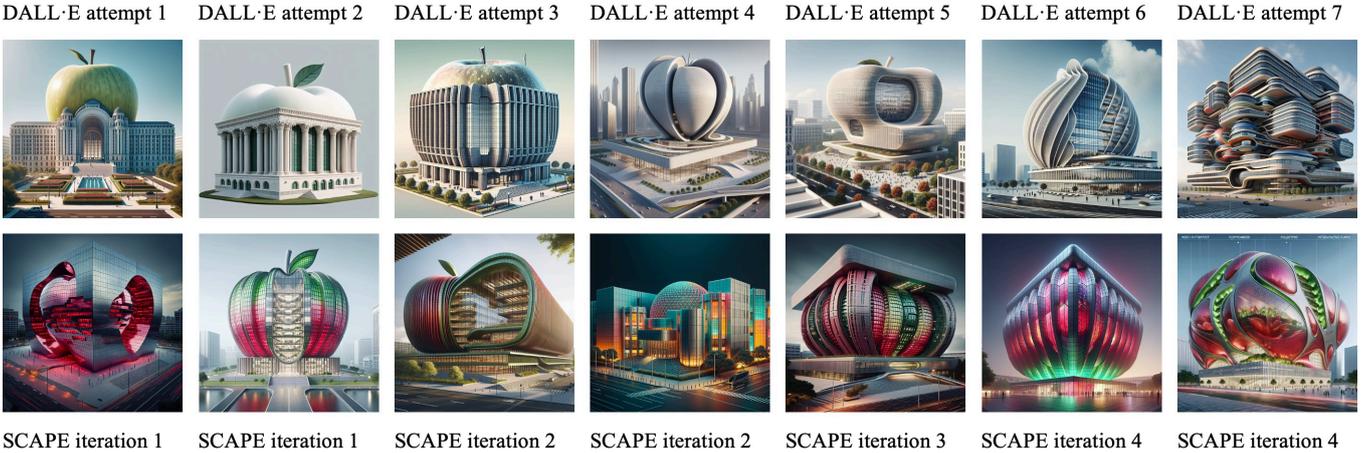

Fig. 9. Results from seven DALL·E 3 prompt attempts (top) vs four SCAPE iterations (bottom), showing parents. (Iteration 3 used a parent from iteration 1.)

Fig. 8 shows an example with the prompt "a government building shaped like a big apple" to illustrate. The images show that DALL·E 3 consistently produces architecturally unrealistic and naïve merging of an apple with a standard-looking building. SCAPE first generation produces plausible apple-inspired buildings and after exploration can produce surprising and innovative variations.

*3) Effectiveness*

We define the effectiveness of SCAPE in terms of how much easier it makes exploring the design space. In SCAPE, users simply provide a short prompt and then iteratively choose (and optionally rate) designs they prefer. In vanilla DALL·E 3 the only method of exploration is to attempt to write an appropriate prompt via trial and error. As an example, starting with the prompt "a government building shaped like a big apple". Fig. 9 shows the progression of designs produced by DALL·E 3 over six attempts at prompt-rewriting. SCAPE achieves more creative and higher quality results in every iteration. Even after many laborious attempts at rewriting prompts, results for vanilla DALL·E 3 are unsatisfactory and difficult to control, with important aspects desired by the user not preserved. For example, for DALL·E 3 attempt 7 the prompt was extended to add "appropriate color" but the resulting image lost most of the morphological features seen from attempt 6, Fig. 9.

*C. Experiment 2 - How Good is SCAPE at Optimisation*

We contacted architects using the AdvisorNet method to find users [28]. The specific steps are as follows: search for people on LinkedIn with the job titles urban planner or architect. Connect with them. If they connect back, we request for them to try SCAPE and provide their feedback. 152 people were invited to connect; 23 tried the tool, a respectable response rate of 15%.

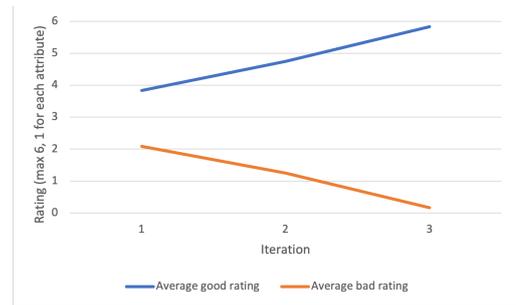

Fig. 10. Change in average numbers of attribute ratings of images against iterations in SCAPE.

In this experiment we are interested in those users who clicked "I'm Done!", selected a favorite image, and gave feedback, thus indicating they had finished exploring and reached a satisfactory result. Fig. 10 shows that the number of attributes rated bad decreases as the number of iterations increase, while the number of attributes rated good increase. (Data for more iterations is sparse as few users continued beyond 3 iterations.) This indicates that SCAPE successfully optimizes as rated by independent architects.

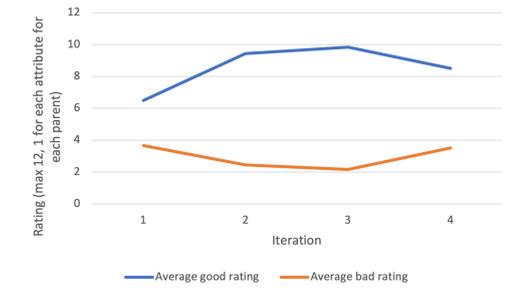

Fig. 11. Change in average numbers of attribute ratings of images against iterations in SCAPE for exploring users.

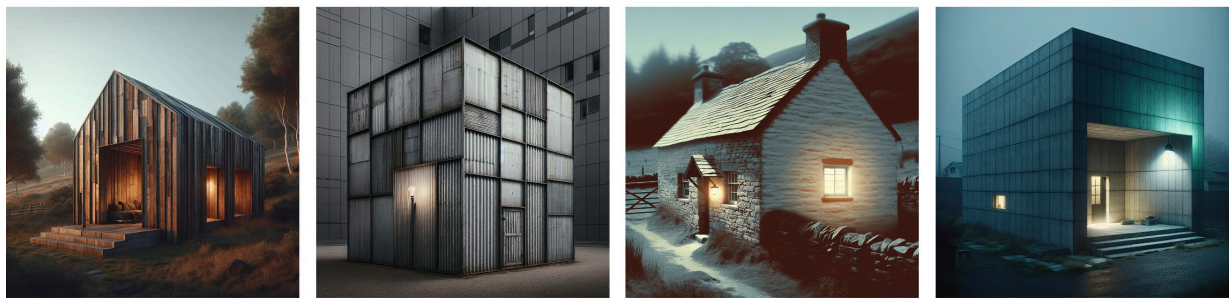
SCAPE iteration 1

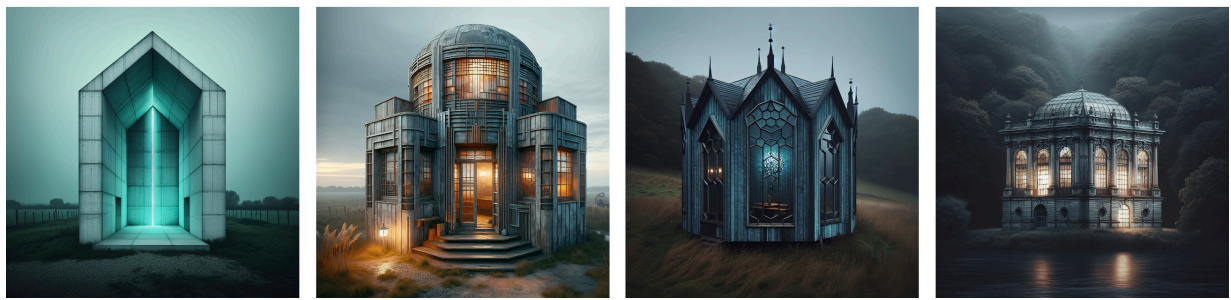
SCAPE final iteration 3

Fig. 12. Example images from first (top) and last (bottom) iterations using prompt "derelict outbuilding with one light on".

TABLE II. POSITIVE FEEDBACK FROM ARCHITECTS

| | |
|---|---|
| 1 | "I've reviewed the AI tool for conceptual architecture and find it quite impressive. Overall, the tool is excellent for architects seeking inspiration. I've obtained some creative results that are not only innovative but also achievable in real-world architecture, rather than being merely unclear or unattainable concepts." |
| 2 | "This is an amazing tool, congratulations on developing it." |
| 3 | "The tool is very easy to use. It is very user-friendly. Selecting what you like and dislike is very flexible!" |
| 4 | "Results and variations are really good, it acts as a mid-journey concept but with a more architectural end product, which is quite helpful for client meetings. the less description will give more ways to generate various options. I would love to use it further." |
| 5 | "It was really cool, surprised at how good it was!!!" |

### D. Experiment 3 - How Good is SCAPE at Exploration

We compare the results from the first generation produced by SCAPE and the final generation, looking at how different the images are. To achieve a consistent and objective comparison we use GPT-4V and ask it to give us a 0-10 score on how different pairs of images are. We run pairwise comparisons between the first and last generation and take the average score, for ten different prompts. The average difference score was 8.001 (SD = 0.306) – a novelty increase of 23.86% from the first iteration, shown in Experiment 1.1, indicating that as iterations progress, SCAPE finds images comprising more and more different concepts as measured by GPT-4V. Thus, SCAPE is able to enable exploration in addition to optimization. Fig. 11 shows average number of good and bad ratings for users that do not click on "I'm done!" indicating more explorative behaviour – good ratings increase and then remain stable. Fig. 12 illustrates the diversity of output possible from SCAPE from the same initial user text.

### E. Experiment 4 - How do Real Architects find the Tool

The overwhelming response from users was positive. Feedback was not always given, but when provided it was constructive and useful. Several suggestions were made on how SCAPE can be improved in the future. Table II provides examples of the positive feedback received and Table III shows feature requests and ways to improve the tool, which we discuss in Section V. Fig. 13 provides examples of images by the architects with the corresponding initial prompts and attribute values generated by GPT-4.

## V. DISCUSSION

The experiments show that an evolutionary algorithm can inject randomness in a controlled manner into the prompts for generative AI, improving its ability to generate novelty, to optimize, and to explore, all without the user needing to rewrite prompts or perform any form of prompt engineering. At present there is a rate limit of five images per minute, meaning that it was necessary to throttle the system to prevent the limit being exceeded by users. Future versions without such a limit would be faster, with the result that users would evolve their concepts for more iterations.

TABLE III. FEATURE REQUESTS AND POINTS TO IMPROVE

| | |
|---|---|
| 1 | "The option to pre-select architectural styles, materials, and other features before generating images would allow architects to tailor the AI outputs more specifically to their project needs." |
| 2 | "The tool was not always accurate in terms of numerical details, such as the count of buildings or floors, If it could be made more sensitive to these details, it would considerably enhance its utility. Addressing this could make the tool perfect for practical applications." |
| 3 | "The text in the image is in some other language and very pixelated." |
| 4 | "It it not clear what architectural style: good or bad means. Because I put in bad referring to the tropical modernism used in the image is bad, but the iteration changed the style to art deco/neo-futurism, etc. So I didnt mean to change, rather to make it look in another way in tropical modernism" |
| 5 | "Is there any way I can go back if I pick the wrong image?" |
| 6 | "It would be nice to have some tip for the descriptions at the beginning, for actually architects do more drawings than writing." |
| 7 | "Is it possible to upload a sketch and translate into CGI? That would be more helpful for the actual work process." |
| 8 | "Further features can be added, which are vital, like number of floors, features like courtyard, landscaping, roof style, etc." |

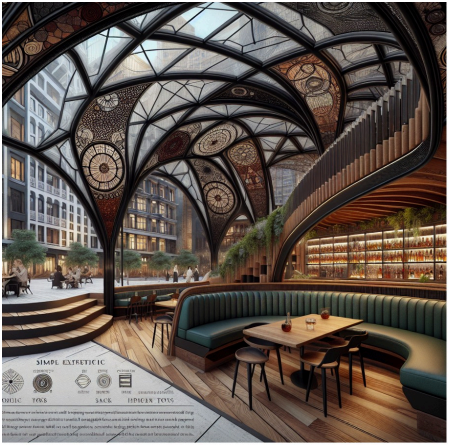 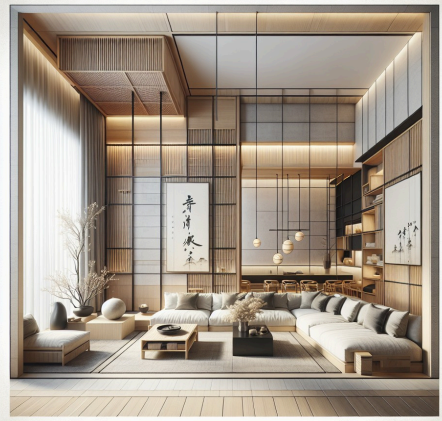 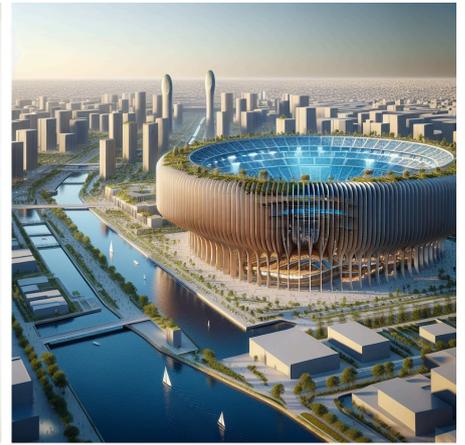

**"Interior design for restaurant themed by source of food seed, black and wood coloured, simple and exotic experience, urban context, cosy feeling, reflect Mediterranean culture"**
'Art Nouveau', 'Black, wood tones', 'Warm dimmed ambient lighting', 'Reclaimed wood, blackened metal', 'Geodesic dome', 'Urban mixed-use development'

**"A living room with a Japanese minimalist and classic combined style"**
'Classic', 'Neutral tones with accents of black and natural wood', 'Warm ambient with decorative chandeliers', 'Natural wood, rice paper, bamboo, stone', 'Clean lines with an emphasis on horizontal elements', 'Urban residence'

**"Stadium that represent the history of waref city and fishing nets industry, blended with city context"**
'Neo-futurism', 'Maritime blue and fishing heritage tan', 'Soft bioluminescent-like fixtures for a natural glow', 'Steel, glass, and recycled fishing nets', 'Vertical garden-clad towers', 'Suburban Greenbelt along the river's edge'

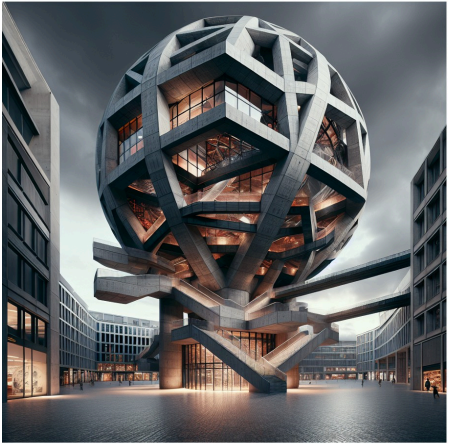 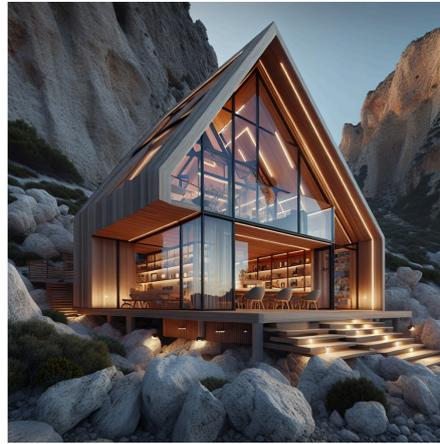 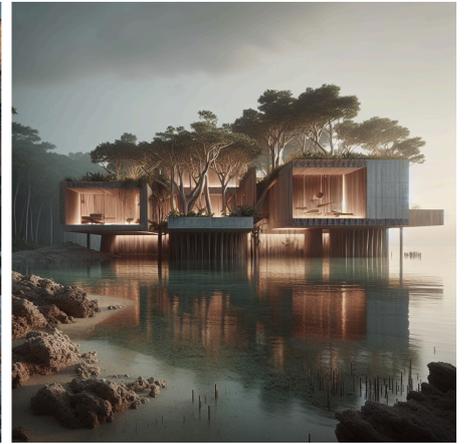

**"A building designed by oma"**
'Deconstructivism', 'Grey and white', 'Halogen ambient lighting', 'Reinforced concrete, wood, glass, and steel fusion', 'Cylindrical column', 'Urban center'

**"A triangular form cottage made of wood with see through glass in the rocky terrain of the sea shore"**
'Contemporary', 'Earthy tones', 'Warm LED lights', 'Timber and transparent glass', 'Geometric', 'Coastal cliffs'

**"Mangrove island landscape luxury resort"**
'Island Vernacular', 'Jungle green, earth brown, and sky blue', 'Diffused lighting with an emphasis on ambiance', 'Reclaimed wood, natural fibers, glass, and local stone', 'Organic shapes with over-water structures', 'Mangrove Forest Island'

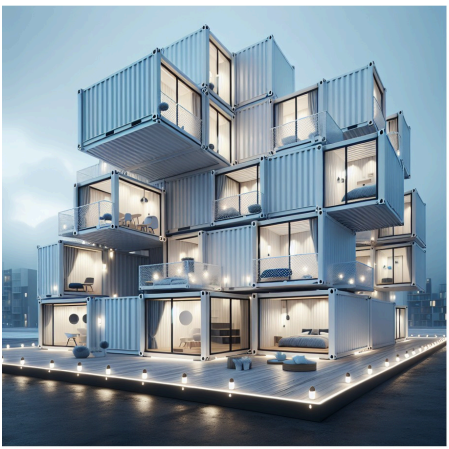 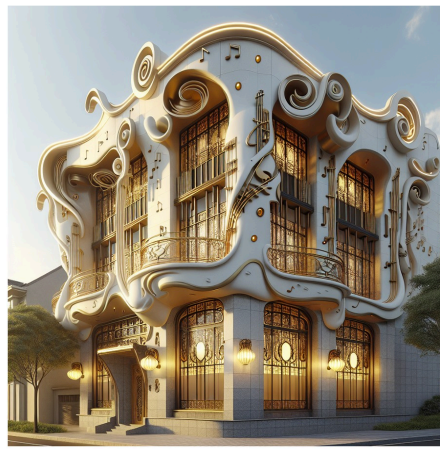 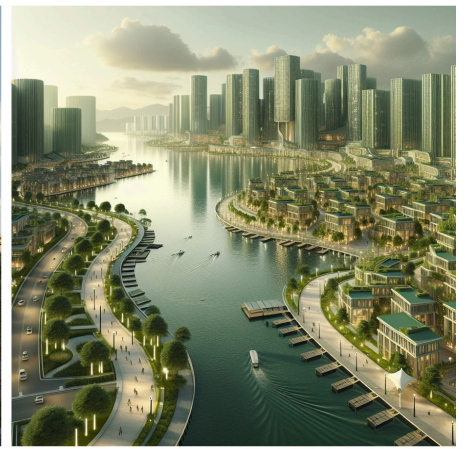

**"Modular container 1 bed housing"**
'Contemporary', 'Monochromatic whites with blue accents', 'Soft LED track lighting', 'Shipping containers and wood', 'Cuboidal-Cylindrical Hybrid', 'Urban high-rise'

**"A musical building"**
'Art Nouveau', 'gold and ivory', 'warm ambient sconces', 'stone and wrought iron', 'organic curves', 'historic neighborhood'

**"A city with more waterways and pedestrian routes, less space for cars."**
'Eco-Friendly', 'Green and beige', 'Solar-powered fixtures', 'Reclaimed timber and recycled plastic', 'Curvilinear', 'Lakeside promenade'

Fig. 13. Examples of images generated with SCAPE by users, showing initial prompts and evolved attributes.

Feedback from users provide useful avenues for future work. Looking at Table III, the first user request asked for the ability to specify attribute values; this is achieved automatically by extracting the values from the textual prompt so the algorithm needs to be made clearer for users. The second and third comments describe problems with DALL·E 3. Numerical limitations of DALL·E 3 are apparent in normal usage (ask for seven buildings and you may only get five or three). We anticipate future versions of DALL·E will likely be better able to count. Likewise, the inability to produce valid text is well-known in generative AI models and is likely to be improved shortly. The fourth comment describes the learning experience of a user as they discover the effect of rating an attribute bad (it mutates to a new value). To achieve what the user desired, they could have either rated this attribute good or left it unrated to preserve that style and see variations. The fifth comment asks if they can backtrack – SCAPE does permit earlier images in past generations to be chosen as parents, so using this approach you can undo accidental selections (or incorporate earlier ideas that are considered useful). Improvements to the user interface and instructions should help make that clearer to users. The sixth comment confirms our reasons for SCAPE as it suggests that architects are less adept at writing prompts – the current input method of DALL·E 3 and most LLM-based generative image AIs. The penultimate comment is an excellent one: supplement the textual description with a sketch. This is becoming possible with some generative AIs and could be a valuable future addition to SCAPE. The final comment suggests the addition of new attributes; a future version of SCAPE could enable users to add their own attributes. These would have their values automatically filled by the LLM using the existing approach. Finally in our own testing and perhaps because of its training data, we observe that the generative AI did not always follow its prompt correctly – "affordable dwelling" or "three-bedroom house" often produced images of large luxury mansions. Such issues could be resolved by finetuning the generative model with specialized architectural training data.

VI. CONCLUSIONS

This paper introduced SCAPE (Searching Conceptual Architecture Prompts using Evolution), a new tool that combines evolutionary search with generative AI to exploit the best aspects of both. We have shown that vanilla generative AI with hand-written prompts lacks the novelty, quality and effectiveness of using human-in-the-loop evolution to help generate prompts. SCAPE ensures quality results are generated by making use of architecturally appropriate attributes (genes) with values automatically derived from the user's input. SCAPE makes use of the language capabilities of GPT to perform crossover and mutation, injecting randomness and hence creativity into a machine learning algorithm designed to produce predictable results. SCAPE provides memoryless DALL·E 3 with a memory in the form of attribute values previously explored and considered good or bad by the user, enabling effective evolutionary search which can both explore and optimize output. When tested by independent architects, the response was markedly positive.